\documentclass{article}
\pdfoutput=1

\PassOptionsToPackage{numbers, square}{natbib}
\usepackage[preprint]{neurips_2024}

\usepackage[utf8]{inputenc} % allow utf-8 input
\usepackage[T1]{fontenc}    % use 8-bit T1 fonts
\usepackage{hyperref}       % hyperlinks
\usepackage{url}            % simple URL typesetting
\usepackage{booktabs}       % professional-quality tables
\usepackage{amsfonts}       % blackboard math symbols
\usepackage{nicefrac}       % compact symbols for 1/2, etc.
\usepackage{microtype}      % microtypography
\usepackage{xcolor}         % colors
\usepackage{graphicx}
\usepackage{hyperref}
\usepackage{authblk}
\usepackage{amsmath}
\usepackage{graphicx}
\usepackage{subcaption}
\usepackage{multirow}
\usepackage{tabularray}
\usepackage{threeparttable}
\usepackage{makecell}
\usepackage{placeins}

\newcommand{\horizon}{H}
\newcommand{\contextlength}{C}
\newcommand{\timeserieslength}{T}
\newcommand{\datasettscount}{M}
\newcommand{\prediction}{\hat{\mathbf{Z}}}
\newcommand{\pearson}{PCC}
\newcommand{\ts}{t}

\title{Evaluating the effectiveness of predicting covariates in LSTM Networks for Time Series Forecasting}
\author{Gareth Davies\\
Neural Aspect \\
\texttt{garethmd@googlemail.com} }
\date{\today}

\begin{document}

\maketitle

\begin{abstract}
Autoregressive Recurrent Neural Networks are widely employed in time-series forecasting tasks,
demonstrating effectiveness in univariate and certain multivariate scenarios. However, their inherent structure
does not readily accommodate the integration of future, time-dependent covariates. A proposed solution, outlined by
Salinas et al 2019\cite{salinas2019deepar}, suggests forecasting both covariates and the target variable
in a multivariate framework. In this study, we conducted comprehensive tests on publicly available time-series datasets, artificially
introducing highly correlated covariates to future time-step values. Our evaluation aimed to assess the
performance of an LSTM network when considering these covariates and compare it against a univariate baseline. As part of this study we
introduce a novel approach using seasonal time segments in combination with an RNN architecture, which is
both simple and extremely effective over long forecast horizons with comparable performance
to many state of the art architectures. Our findings from the results of more than 120 models reveal that under certain conditions
jointly training covariates with target variables can improve overall performance of the model, but often there exists a significant performance
disparity between multivariate and univariate predictions. Surprisingly, even when provided with covariates informing the
network about future target values, multivariate predictions exhibited inferior performance.
In essence, compelling the network to predict multiple values can prove detrimental to model performance, even in the presence of informative covariates.
These results suggest that LSTM architectures may not be suitable for forecasting tasks where predicting covariates
would typically be expected to enhance model accuracy.
\end{abstract}

\section{Introduction}
Forecasting future events, is a critical endeavour across various domains, facilitating informed decision-making
and resource allocation. These forecasting tasks heavily rely on historical data to make accurate predictions. Given the inherent 
sequential and time-dependent nature of such data, Recurrent Neural Networks (RNNs) emerge as a natural choice for modelling temporal sequences 
due to their ability to retain memory across time steps.\cite{elman1990}

However, traditional forecasting methods, particularly autoregression, which relies on using past observations to 
predict future values, encounter challenges as forecasting horizons extend. As predictions are recursively 
dependent on preceding values, errors can compound over time, resulting in diminished accuracy.

Furthermore, predicting values solely from prior observations prevents analysis of the underlying features that influence future 
outcomes. Therefore making it impossible to simulate scenarios or identify corrective action. For example during the Covid 
pandemic the number of positive tests informed future hospitalisation and mortality rates, or the volume of sales 
meetings could inform an organisation of future sales.  Clearly the inclusion of these leading indicators would be highly desirable 
both in terms of improving accuracy and our understanding of the domain.

To address these limitations and enhance prediction accuracy, researchers have explored the incorporation of 
additional information, known as covariates, into forecasting models. Covariates can provide valuable context 
and assist the network in making more informed predictions. They can be either static time-independent, which are fixed 
for an entire time-series, or time-dependent, where provided values can change at each time step. 
When future time-dependent covariates are known in advance, such as week or month numbers, they can be 
leveraged to augment autoregressive predictions. Conversely, in scenarios where future covariates are unknown, 
they must either be estimated beforehand or predicted simultaneously with the target variable, rendering each 
prediction multivariate.

In this study, we aim to evaluate the efficacy of employing autoregressive multivariate Long Short-Term Memory 
(LSTM) models compared to traditional univariate models without covariates. We conduct our investigation 
within the context of well-established time series datasets from the Monash repository \cite{DBLP:conf/nips/GodahewaBWHM21}, 
a widely recognised benchmark for evaluating forecasting models. Our methodology involves constructing baseline univariate models 
using an autoregressive LSTM, ensuring their performance aligns with established benchmarks. Subsequently, we 
augment each dataset by introducing covariates engineered from future time-step values, effectively guiding 
the network in predicting both the target variable and the associated covariate. We then assess the performance 
of our models based on predicting target values over a forecast horizon $\horizon$.

The contributions of this study are:
1. A novel approach to modelling timeseries data with LSTM networks that is extremely effective in a univariate setting and does not 
require specialised feature engineering.
2. Quantification of the effect cross correlation between covariates and target variables has on the resulting performance of a trained model.
3. An evaluation of how model performance is affected as forecast horizons increase. 
4. A simple approach to synthesise covariates from univariate datasets for running experiments in a controlled and repeatable manner.

\section{Related Work}
The use of covariates in State Space models have been studied notably by Wang\cite{wang2006} and more recently by Puindi and Silba 
\cite{puindi2020dynamic}. Wang proposed ESCov a method that extended HoltWinters to utilise informative covariates which were calculated in advance 
of the main modelling task. Wang observed that the inclusion of covariates could improve the overall accuracy of the model, the degree to which is in part
determined by the quality of the predictions of the covariates. 

Lai et al. \cite{lai2018modeling} developed LSTNet, which supports multivariate scenarios and which combines a CNN and a GRU to capture both 
long and short-term dependencies over time, as well as local dependencies between variables. 

Salinas et al. \cite{salinas2019deepar} proposed DeepAR, an autoregressive LSTM model intended to be used specifically 
for time-series forecasting. Their approach included time dependent covariates as inputs into each timestep combining them 
with the outputs of the previous timestep to assist the network.

Salinas et al subsequently extended DeepAR to a multivariate setting with DeepVAR and GPVAR \cite{salinas2019highdimensional}, which supported 
scenarios where datasets of multiple timeseries are forecasted simultaneously through vector autoregression. The objective being to model the 
dynamics between the target variables in each timeseries. As the dimension of the output vector scales, the multivariate predictions can 
become very high dimensional, and encounter high computational costs as a result.  GPVAR addresses this by modifying the output model 
with a low rank covariance structure. 

More recently, transformer-based models such as Informer, Autoformer, and Crossformer
\cite{zhou2021informer,wu2022autoformer, zhang2023crossformer} have emerged for multivariate predictions. Notably Informer used a 
Weather, WTH,  which could be configured to use in a covariate setting. These models extract feature maps from the temporal dimension and extend the 
vanilla Transformer attention mechanism to produce more accurate predictions over longer forecast horizons. 
A study by Zeng et al \cite{zeng2022transformers} found that the inclusion of covariates in transformer networks were largely ineffective, proposing instead 
simple linear models with seasonal decomposition similar to traditional state space ETS models.

Chen et al, proposed TSMixer \cite{chen2023tsmixer}, a FCC architecture which reshapes the inputs to force the network to learn across feature and temporal dimension in an 
attempt to improve multivariate performance.

\section{Methodology}
In this section, we outline our methodology for assessing the effectiveness of LSTM networks in 
multivariate time series prediction tasks, particularly focusing on the impact of time-dependent covariates.

\subsection{Problem Statement}
We consider a dataset containing $\datasettscount$ timeseries where each timeseries contains $\timeserieslength$ historical observations, $\mathbf{y}$ 
and covariates $\mathbf{X}$ with $k$ dimensions or individual features such that:
\begin{align*}
\mathbf{y} &= [y_1, y_2, \dots, y_T] \\
\mathbf{X} &= \begin{bmatrix}
x_1^{(1)} & x_1^{(2)} & \ldots & x_1^{(k)} \\
x_2^{(1)} & x_2^{(2)} & \ldots & x_2^{(k)} \\
\vdots & \vdots & \ddots & \vdots \\
x_T^{(1)} & x_T^{(2)} & \ldots & x_T^{(k)}
\end{bmatrix}
\end{align*}
Let $\mathbf{Z} = \mathbf{y} \oplus \mathbf{X}$. The objective of the training task is to forecast a vector of future values and covariates across a forecast horizon $\horizon$. 
At each timestep $t$, this vector is represented as ${\mathbf{\hat{z}}_t} = [\hat{y_t}, \hat{x_t}^1, \hat{x_t}^2, \dots, \hat{x_t}^k]$ where $\hat{y_t}$ is the predicted target variable
and $\hat{x_t}^1, \hat{x_t}^2, \dots, \hat{x_t}^k$ are the predicted future covariates. 

During model evaluation, we derive the vector of predicted target variables, $\mathbf{\hat{y}}$ from $\prediction$ such that $\mathbf{\hat{y}} = [\hat{y_1}, \hat{y_2}, \dots, \hat{y_\horizon}]$.
The aim here is to generate $\mathbf{\hat{y}}$ in a manner that ensures it's more accurate than what could be achieved solely by relying on historical observations.

\subsection{Datasets}
We selected four publicly available datasets from the Monash repository \cite{DBLP:conf/nips/GodahewaBWHM21} 
namely \texttt{Hospital} \cite{godahewa_2021_4656014}, \texttt{Tourism} \cite{godahewa_2021_4656096}, \texttt{Traffic} \cite{godahewa_2021_4656135} 
and \texttt{Electricity} \cite{godahewa_2021_4656140}. These datasets are from diverse domains and have 
been extensively benchmarked in prior research by Godahewa et al \cite{DBLP:conf/nips/GodahewaBWHM21} and are commonly used for evaluating time series 
forecasting models. The \texttt{Hospital} dataset consists of 767 monthly time series depicting patient counts related to 
medical products from January 2000 to December 2006. 
Similarly, the \texttt{Electricity} dataset was used by Lai \cite{lai2018modeling} and represents the hourly electricity 
consumption of 321 clients from 2012 to 2014. The \texttt{Tourism} dataset comprises monthly figures from 1311 tourism-related time series, 
while the \texttt{Traffic} dataset includes 862 weekly time series showing road occupancy rates on San 
Francisco Bay area freeways from 2015 to 2016. Each dataset is accompanied by specific context 
lengths $\contextlength$ (lookback window) and forecast horizons $\horizon$, providing a robust evaluation framework.

\subsection{Covariate Data Augmentation}
Since the original datasets do not contain time-dependent covariates, we artificially augment them 
to introduce covariates correlated with future target values such that $x_t^k$ is correlated with $y_{t+k}$.
When k=3, for example, the network is supplied with 3 covariates correlated with the following 3 target values. 
The network would have to learn that the value of the covariate, $x_t^k$, would take effect on target, $y$, at a time $k$ timesteps in the future. 

Noise is added to each leading indicator to control the level of cross correlation between the covariate and its 
corresponding target. The noise is calculated as follows: 

\begin{itemize}
\item For each time series in the dataset the mean $\mu$ and standard deviation $\sigma$ of $y$ are computed.
\item For each covariate value a sample $\epsilon$ is drawn from a unit normal distribution; $\epsilon \sim N(0,1)$ 
\item The noise is computed by scaling the random $\epsilon$ values by $\mu$ and $\sigma$ and then further
scaled by an error level factor $\gamma$:
$\gamma \in \{0, 0.1, ..., 1.9\}$
\end{itemize}
Let $\mathbf{x}^{(k)}$ be the augmented covariate input sequence $[x_0, x_1, ..., x_T]$, 
then $x^{k}_t = y_{t+k} + \gamma \cdot \mu \cdot \epsilon + \gamma \cdot \sigma \cdot \epsilon $
\begin{figure}[ht]
\centering
\includegraphics[width=0.8\textwidth]{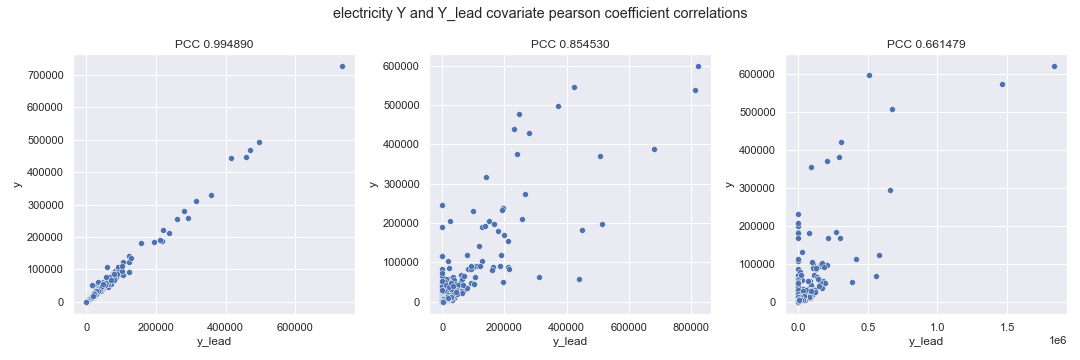}
\caption{Plots of various correlations between covariate and target variables on the Electricity dataset}
\label{fig:my_label}
\end{figure}
We then compute the cross correlation between the covariates and the target variables with the Pearson correlation coefficient ($\pearson$) 
and aim for each experiment to be run with a strong to perfect positive correlation in the range 0.5 to 1.0. This value is then directly 
comparable to autocorrelation of lagged variables used in techniques such as ARIMA as described by Hyndman et al \cite{hyndman2021forecasting}.

\subsection{Model architectures}
To better understand how covariates affect our results and attribute their impact more accurately, 
we employed two straightforward LSTM based architectures. Unlike the deep learning models benchmarked 
by Monash \cite{DBLP:conf/nips/GodahewaBWHM21} which used the GluonTS \cite{gluonts_arxiv} framework 
for training, we did not add covariates such as lag variables and static time-series identifiers in order to 
accurately evaluate the true effect of the augmented leading covariates. 

Of these models: base-lstm is used as a baseline and is benchmarked against the Monash repository results using the same context length $\contextlength$
and forecast horizons $\horizon$; while the seg-lstm model is intended to be more capable of forecasting over longer horizons using longer context lengths.

\subsubsection{Baseline model (base-lstm)}
Baseline is a vanilla LSTM with a scaling strategy inspired by DeepAR\cite{salinas2019deepar}.
The input data is scaled using mean absolute scaling within each mini-batch. An additional time dependent feature $\log(scale)$ is concatenated with the input vector. The scaled values are fed into a $2$ layer 
LSTM with $40$ cells in each layer. The output of the LSTM is then fed into a single fully connected layer 
with $40$ cells followed by ReLU activations. Finally, an output layer generates a vector $\mathbf{z_t}$ 
containing the predicted target variable $\hat{y_t}$ and predicted covariates $\mathbf{\hat{x_t}}$.  An inverse scaling operation finally transforms 
the predictions back into the original scale.

\subsubsection{Segment model (seg-lstm)}
The seg-lstm model addresses the limitation of the baseline model in forecasting over longer time horizons. 
Our approach aims to maximise historical data without relying on extra features, such as lag variables that get applied 
with GluonTS \cite{gluonts_arxiv} models and is typical with transformer based approaches \cite{zhou2021informer,wu2022autoformer,zhang2023crossformer,kitaev2020reformer}.

To achieve this, we propose an LSTM with a simple modification to handle input data. We reshape each context window of $\contextlength$ values 
into a vector of dimension $\frac{\contextlength}{d} \times d$, where $d$ represents the segment length, typically set to the seasonality
 of the data (e.g., 24 for hourly readings).

The model architecture is similar to the base-lstm, consisting of a straightforward LSTM with 2 layers and 40 cells in each layer, followed by fully connected 
layers with ReLU activations. We use mean absolute scaling but omit the additional time-dependent feature $\log(\text{scale})$.

During training, we compute the loss by comparing the last timestep of each segment output to the corresponding timestep in the 
target vector shifted one step into the future. The network is trained to predict one timestep ahead from each segment, with hidden 
state representations derived from segments are separated by a time period $d$.

During inference, we predict the next timestep vector, append it to the previous timestep segment, and drop the oldest vector $\mathbf{z}_{t_0}$. 
This forms an autoregressive loop, predicting successive segments of adjacent timesteps rather than spanning a fixed time period $d$.

\subsection{Evaluation Metrics}
We evaluate the performance of each model using MAE, RMSE, and sMAPE, computed on the raw (i.e. unscaled) 
values of the dataset. This approach ensures direct comparability with results from 
the Monash archive\cite{DBLP:conf/nips/GodahewaBWHM21}, enabling comprehensive assessment and comparison.
\begin{equation*}
\begin{aligned}
\text{MAE}(\hat{Y}, Y) &= \frac{ \sum_{\ts=1}^{\horizon}|\hat{Y}_\ts - Y_\ts|}{\horizon} &
\quad
\text{RMSE}(\hat{Y}, Y) &= \sqrt{\frac{\sum_{\ts=1}^{\horizon}(\hat{Y}_\ts - Y_\ts)^2}{\horizon}} \\
\text{sMAPE}(\hat{Y}, Y) &= \frac{100\%}{\horizon}\sum_{\ts=1}^{\horizon}\frac{|\hat{Y}_\ts - Y_\ts|}{(|\hat{Y}_\ts| + |Y_\ts|)/2}
\end{aligned}
\end{equation*}
\section{Experimental Setup}
We now describe the details of the experiments including how the data was processed, the 
model training and evaluation.

\subsection{Data Processing}
To facilitate model training, we generate time windows of data, each of which is equal in 
size to the sum of the context length ($\contextlength$) and the forecast horizon ($\horizon$). During each training epoch, 
complete and overlapping windows are extracted from each time series. 

The dataset is partitioned into training, validation, and test sets, following a chronological order. 
The validation set excludes values within the last forecast horizon, while the training set excludes 
values from the last two forecast horizons. Evaluation for both validation and testing is conducted 
on the last complete forecast horizon period of each time series. This approach ensures direct comparison to 
the Monash benchmarks \cite{DBLP:conf/nips/GodahewaBWHM21} and maximises the 
utilisation of training data while preventing data leakage between the three sets.

We employ a variation of mean absolute scaling to standardise the data within each mini-batch (local normalisation), 
accounting for batches with values equal to zero.
\begin{equation}
\text{scale}(\mathbf{z}) = 
\text{max}( \frac{ \sum_{\ts=1}^{\contextlength}|z^{(y)}_\ts|}{\contextlength},  1)
\end{equation}
\subsection{Training}
We utilised teacher forcing which trains the network to predict a single
timestep ahead and calculated the loss for the entire sequence (ie the context length and the forecast horizon).
The network was trained as a regression task with a Smooth F1 Loss objective.
\begin{equation}
\text{SmoothL1Loss}(\hat{z}, z) = 
\begin{cases} 
0.5(\hat{z} - z)^2 & \text{if } |\hat{z} - z| < 1 \\
|\hat{z} - z| - 0.5 & \text{otherwise}
\end{cases}
\end{equation}
A number of hyperparameters were set to match the default
GluonTS parameters \cite{gluonts_arxiv} including weight decay of $1e-8$; dropout  of 0.1; and a learning rate of 0.001. 
We used OneCycle scheduling to anneal and then 
decrease the learning rate\cite{smith2018superconvergence} over 100 epochs. The models were optimised using AdamW. We made no attempt to optimise hyperparameters. 

Performance was measured on the validation set at the end of each epoch by measuring the Smooth L1 loss using
free running (i.e. using the predicted outputs of one timestep as the input into the next). The model checkpoint
with the lowest validation loss was selected for testing. The hidden state of the LSTM was initialised to $0$. 

\subsection{Experimental Procedure}
Initially, we trained base-lstm models for each dataset using the context length and forecast horizon values consistent with 
those used in the Monash repository which allowed us to compare the performance of our models 
against benchmarked results. 

Next, we trained and evaluated models under various scenarios involving 1, 2 and 3 
covariates. For the number of covariates in the scenario a set of base-lstm models were trained with increased noise to vary the 
level of cross correlation with the target variables.  This allows us to assess the impact of covariates on model performance under 
different correlation levels. These models were trained to a context length and forecast horizon that equalled the benchmarked 
results from Monash \cite{DBLP:conf/nips/GodahewaBWHM21}

We then trained seg-lstm models using the same procedure with the exception that we use a longer context 
length $\contextlength$ which was set to 3x the forecast horizon $\horizon$ on \texttt{Hospital}, \texttt{Tourism} and \texttt{Electricity} and 8x the forecast horizon on \texttt{Traffic}. 

\section{Results}
Table \ref{tab:benchmark} presents the mean error metrics for each dataset evaluated against the neural network architectures benchmarked 
by Godahewa et al \cite{DBLP:conf/nips/GodahewaBWHM21}. Among these architectures, FFNN and N-BEATS \cite{oreshkin2020nbeats} are fully-connected models, 
while DeepAR \cite{salinas2019deepar} is based on an LSTM network. WaveNet \cite{oord2016wavenet}, originally designed for 
audio synthesis, was adapted by Alexandrov et al \cite{gluonts_arxiv} for time-series tasks in GluonTS. Additionally, the Transformer 
architecture closely follows the implementation described in the original paper by Vaswani et al \cite{vaswani2023attention}.

Comparing the base-lstm model across datasets, it yielded comparable results on the \texttt{Hospital} and \texttt{Traffic} datasets, 
slightly inferior results on \texttt{Tourism}, and significantly poorer results on \texttt{Electricity} when measured by sMAPE. However, 
the performance is marginally better when evaluated using MAE and RMSE. Notably, the base-lstm model exhibited relatively better performance on datasets with shorter forecast horizons.

The seg-lstm model emerged as the top performer on Electricity with RMSE and second with MAE and sMAPE, whilst on \texttt{Tourism} it ranked second  
with MAE and RMSE , both of which have longer forecast horizons. Overall, both of our models demonstrated comparable performance to the benchmarks on the shorter forecast 
horizons of \texttt{Hospital} and \texttt{Traffic}, while the seg-lstm model also displayed competitive performance on \texttt{Tourism} 
and \texttt{Electricity}.
\begin{table}[tbp]
  \caption{Benchmark results for each dataset}
  \centering
  \begin{threeparttable}
  \begin{small}
  \renewcommand{\multirowsetup}{\centering}
  \setlength{\tabcolsep}{2.6pt}
  \begin{tabular}{c|c|ccccccc}
    \toprule
    \multicolumn{2}{c}{Models} & \multicolumn{1}{c}{FFNN} &  \multicolumn{1}{c}{DeepAR} & \multicolumn{1}{c}{N-Beats}  & \multicolumn{1}{c}{Wavenet} & \multicolumn{1}{c}{Transformer} & \multicolumn{1}{c}{base-lstm*} & \multicolumn{1}{c}{seg-lstm*}  \\
    \toprule
    \multirow{4}{*}{\rotatebox{90}{Hospital}}
    \multirow{4}{*}{{(12)}} 
    & sMAPE &  18.33         & \textbf{17.45}  & 17.77            & 17.55            & 20.08                              & 17.52                           & 18.05                     \\
    & MAE   & 22.86         & 18.25            & 20.18            & 19.35            & 36.19                              & \textbf{18.03}                  & 19.95                     \\
    & RMSE &  27.77         & \textbf{22.01}   & 24.18            & 23.38            & 40.48                              & 22.03                           & 24.19                     \\
    \\
    \midrule
    \multirow{4}{*}{\rotatebox{90}{Tourism}}
    \multirow{4}{*}{{(24)}} 
    & sMAPE & 20.11         & \textbf{18.35}   & 20.42            & 18.92            & 19.75                              & 21.50                           & 19.85                     \\
    & MAE   & 2022.21       & \textbf{1871.69} & 2003.02          & 2095.13          & 2146.98                            & 2336.42                         & 1956.07                   \\
    & RMSE  & 2584.10       & \textbf{2359.87} & 2596.21          & 2694.22          & 2660.06                            & 2964.96                         & 2413.64                   \\
    \\
    \midrule
    \multirow{4}{*}{\rotatebox{90}{Traffic}} 
    \multirow{4}{*}{{ (8)}} 
    & sMAPE & 12.73         & 13.22            & 1\textbf{2.40}   & 13.30            & 15.28                              & 12.77                           & 12.97                     \\
    & MAE   & 1.15          & 1.18             & \textbf{1.11}    & 1.20             & 1.42                               & 1.15                            & 1.17                      \\
    & RMSE  & 1.55          & 1.51             & \textbf{1.44}    & 1.61             & 1.94                               &  1.56                           & 1.58                      \\
    \\
    \midrule
    \multirow{4}{*}{\rotatebox{90}{Electricity}}  
    \multirow{4}{*}{{(168)}} 
    & sMAPE & 23.06         & \textbf{20.96}   & 23.39            & -                & 24.18                              & 34.12                           & 21.20                     \\
    & MAE   & 354.39        & 329.75           & 350.37           & \textbf{286.56}  & 398.80                             & 525.50                          & 287.95                    \\
    & RMSE  & 519.06        & 477.99           & 510.91           & 489.91           & 514.68                             & 675.03                          & \textbf{469.07}           \\
    \\
    \bottomrule
  \end{tabular}
  \begin{tablenotes}
    \item[*] Mean values from 5 runs.
    \item Forecast horizons are given in the brackets
  \end{tablenotes}
\end{small}
\end{threeparttable}
\vspace{-15pt}
\label{tab:benchmark}
\end{table}

\subsection{Covariates}
The results of covariates compared to univariate cases are presented in table \ref{tab:covariate_results}. For brevity we have included $\pearson$ values of 1, 0.9 and 0.5 (See appendix for details of complete 
results) and show the sMAPE for each dataset at the full forecast horizons as measured in the benchmark tests. We present the results for each value of k covariates tested (1, 2 and 3) and 
compare it to the univariate case where k=0. To some extent the relative differences between the covariate and univariate cases are somewhat specific to the 
dataset. The models trained on \texttt{Traffic} appear to benefit the most from covariates giving improved results in 14 of the 18 experiments, which is perhaps to be expected 
as this has the shortest forecast horizon. Conversely models performed worst on, \texttt{Tourism} only outperforming univariate in 5 scenarios, and that was with perfectly correlated covariates.
\text{Electricity}, which has the longest forecast horizon of 168, marginally outperformed univariate in both models with $\pearson$ = 1.0 using 1 and 2 covariates. 

\begin{table}[tbp]
  \caption{sMAPE results for covariates $k \in \{0, 1, 2, 3\}$ and cross correlation $\pearson \in \{1, 0.9, 0.5\}$}
  \centering
  \begin{threeparttable}
  \begin{small}
  \renewcommand{\multirowsetup}{\centering}
  \setlength{\tabcolsep}{1.8pt}
  \begin{tabular}{c|c|cccccccccccccc}
    \toprule
    \multicolumn{2}{c}{Models} & \multicolumn{3}{c}{\textbf{base-lstm}} & \multicolumn{3}{c}{\textbf{seg-lstm}} \\
    \cmidrule(lr){3-5} \cmidrule(lr){6-8}
    \multicolumn{2}{c}{$\pearson$} & 1 & 0.9 & 0.5 & 1 & 0.9 & 0.5 \\
    \toprule
    \multirow{4}{*}{\makecell{\textbf{Hospital}\\(12)}} & k=0 & 17.52 $\pm$ 0.041 & 17.52 $\pm$ 0.041 & 17.52 $\pm$ 0.041 & 18.05 $\pm$ 0.135 & 18.05 $\pm$ 0.135 & 18.05 $\pm$ 0.135 \\
    & k=1 & 16.13 & 17.45 & 17.77 & 16.07 & 17.96 & 18.09 \\
    & k=2 & 14.94 & 17.65 & 18.06 & 17.33 & 18.49 & 18.51 \\
    & k=3 & 13.49 & 17.71 & 17.72 & 17.71 & 17.68 & 18.77 \\
    \midrule
    \multirow{4}{*}{\makecell{\textbf{Tourism}\\(24)}} & k=0 & 21.50 $\pm$ 0.531 & 21.50 $\pm$ 0.531 & 21.50 $\pm$ 0.531 & 19.85 $\pm$ 0.62 & 19.85 $\pm$ 0.62 & 19.85 $\pm$ 0.62 \\
    & k=1 & 20.54 & 22.17 & 28.18 & 19.14 & 20.32 & 21.56 \\
    & k=2 & 20.26 & 24.33 & 27.64 & 19.07 & 20.49 & 21.84 \\
    & k=3 & 23.08 & 27.59 & 28.13 & 18.61 & 19.96 & 23.06 \\
    \midrule
    \multirow{4}{*}{\makecell{\textbf{Traffic}\\(8)}} & k=0 & 12.77 $\pm$ 0.065 & 12.77 $\pm$ 0.065 & 12.77 $\pm$ 0.065 & 12.97 $\pm$ 0.108 & 12.97 $\pm$ 0.108 & 12.97 $\pm$ 0.108 \\
    & k=1 & 11.36 & 12.59 & 12.89 & 11.64 & 12.85 & 13.00 \\
    & k=2 & 10.73 & 12.15 & 12.62 & 11.10 & 13.12 & 12.86 \\
    & k=3 & 9.57 & 11.86 & 13.51 & 9.88 & 11.94 & 12.90 \\
    \midrule
    \multirow{4}{*}{\makecell{\textbf{Electricity}\\(168)}} & k=0 & 34.12 $\pm$ 2.38 & 34.12 $\pm$ 2.38 & 34.12 $\pm$ 2.38 & 21.20 $\pm$ 0.232 & 21.20 $\pm$ 0.232 & 21.20 $\pm$ 0.232 \\
    & k=1 & 33.10 & 30.41 & 33.95 & 21.01 & 22.37 & 21.45 \\
    & k=2 & 33.62 & 30.42 & 38.16 & 21.14 & 22.48 & 22.11 \\
    & k=3 & 36.18 & 33.83 & 31.50 & 21.61 & 23.04 & 22.20 \\
    \bottomrule
  \end{tabular}
  \begin{tablenotes}
    \item Forecast horizons given in the brackets. Univariate k=0 intervals are 95\% CI.
  \end{tablenotes}
  \end{small}
  \end{threeparttable}
  \label{tab:covariate_results}
  \vspace{-15pt}
\end{table}

\subsubsection{Comparing PCC at forecast horizons = k}
\begin{figure}[ht]
\centering
\includegraphics[width=0.9\textwidth]{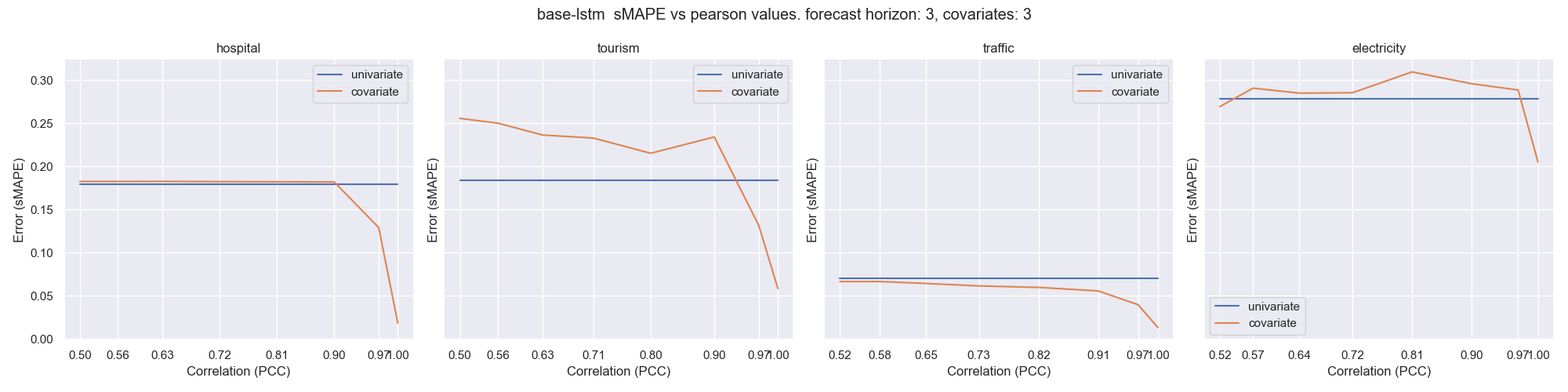}
\caption{base-lstm smape as a function of PCC for 3 covariates}
\label{fig:base_lstm_k_3_smape_vs_pearson}
\end{figure}

We are interested in analysing the performance of models in the special cases where the forecast horizon matches the number of covariates, as in these cases the model 
is effectively provided with information required to predict each timestep. Specifically, we wanted to understand the role that the cross correlation plays 
on model performance. Figure \ref{fig:base_lstm_k_3_smape_vs_pearson} shows how the base-lstm model sMAPE degrades as a function of $\pearson$ on each of the datasets where the 
forecast horizon is 3. Models with 1 and 2 covariates exhibit similar traits, see Appendix. Unsurprisingly the performance is best when the correlation 
is perfect (ie PCC = 1.0) to the extent that an almost perfect prediction was obtained in the cases of \texttt{Traffic} and \texttt{Hospital}. However the performance degrades as the PCC decreases 
in a non linear manner, with the error increasing rapidly as the PCC falls to 0.9. Interestingly even in this highly artificial and advantageous setting, the error on \texttt{Tourism} 
and \texttt{Electricity} are still worse than the univariate case at correlation levels less than 0.9.
\subsubsection{Comparing PCC over longer forecast horizon trajectories}
With the exception of \texttt{Traffic}, as the forecast horizon extends any performance advantage over a univariate setting diminishes for 
a $\pearson$ value of 0.9 and below when forecast horizon exceeds 3-4 timesteps. Figure \ref{fig:tourism_base_vs_seg} plots the base-lstm sMAPE as 
a function of forecast horizon on the \texttt{Tourism} dataset.
\subsubsection{Comparing base-lstm and seg-lstm performance}
\begin{figure}[ht]
  \centering
  \includegraphics[width=1.\textwidth]{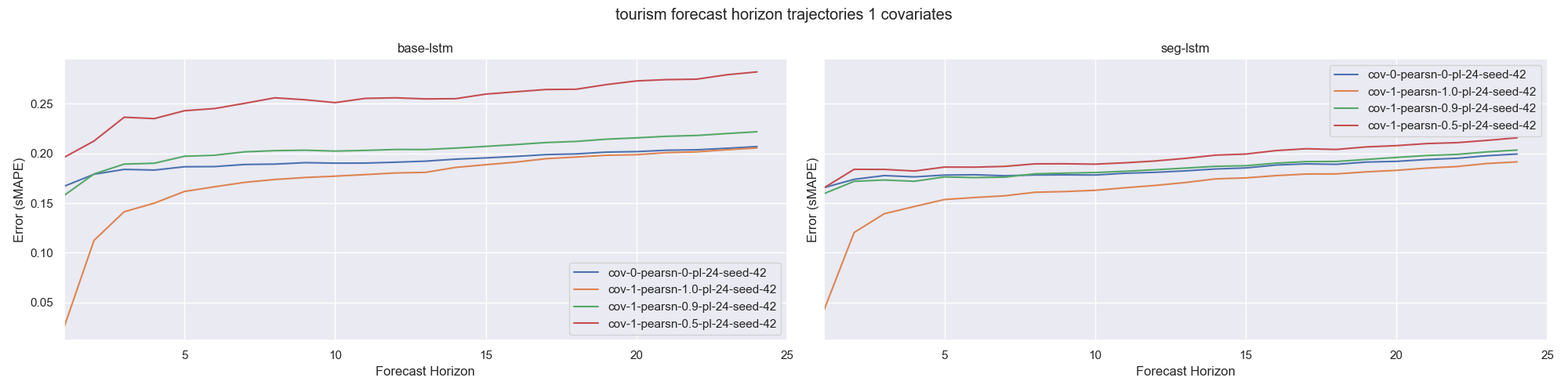}
  \caption{\texttt{Tourism} smape for 1 covariate for base-lstm and seg-lstm}
  \label{fig:tourism_base_vs_seg}
  \end{figure}

Looking at the differences between base-lstm and seg-lstm. We see that both models share common characteristics when examining the errors on forecast horizon trajectories. 
The stronger model in univariate scenarios will also perform better in a covariate setting. Figure \ref{fig:tourism_base_vs_seg} shows how the sMAPE accumulates over the full forecast horizon on 
the \texttt{Tourism} dataset for both base-lstm and seg-lstm. Note how the variance of errors is smaller across the range of PCC values on 
the stronger seg-lstm model, indicating that the additional noise added to lower PCC covariates combines with the model's inherent performance limitations to compound errors over time.
\subsubsection{Comparing covariates across Forecast Horizon trajectories}
\begin{figure}[ht]
\centering
\includegraphics[width=1.\textwidth]{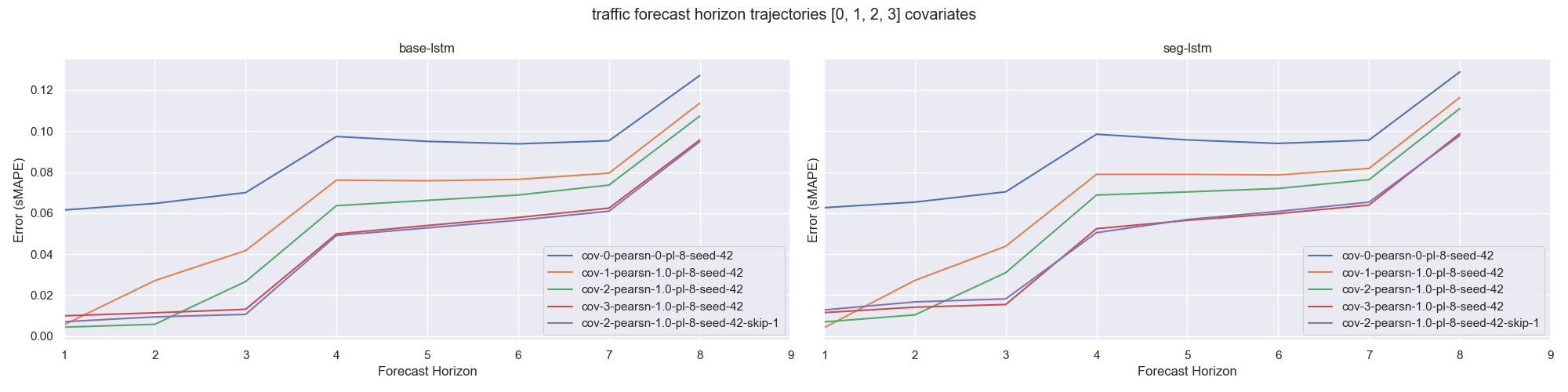}
\caption{\texttt{Traffic} smape comparing univariate to 1, 2 and 3 covariates for base-lstm at a $\pearson$ = 0.9}
\label{fig:lstm_traffic_univariate_vs_covariate}
\end{figure}

Turning to comparing multiple covariates. Fig \ref{fig:lstm_traffic_univariate_vs_covariate} shows sMAPE across full forecast horizons for 1, 2 and 3 covariates on 
the \texttt{Traffic} dataset with the special case of perfect correlation (PCC = 1). Note that the error reduces as the number of covariates increases. 
Interestingly, the addition of each covariate shifts the onset of significant errors by one timestep. (i.e. errors start to accumulate at t=1 for k=1, 
at t=2 for k=2 and t=3 for k=3). We observe identical patterns on the other datasets tested.
Clearly the network is effectively utilising the covariates to learn the values for the first k timesteps, however it is conceivable that a model may be using covariate 
values from the current timestep alone rather than utilising covariates from previous timesteps. For example in the case of 3 covariates  ${x_t^1, x_t^2, x_t^3}$ 
are leading indicators for target values ${y_t, y_{t+1}, y_{t+2}}$, and therefore the network does not require covariates from previous timesteps in order to predict the subsequent 
3 timesteps to produce the error improvement that we observe. Consequently we don't know if the model is utilising covariates across the temporal dimension 
or if it is just using the current timestep's covariates.

One way to help answer this is to omit the $x_t^2$ covariate meaning that at any given timestep the input vector contains lead indicators for $y_t$ and $y_{t+2}$ and the model 
would have to obtain the leading covariate for $y_{t+1}$ from $x^3_{t-1}$ (ie the previous timestep). If the error trajectory in this example continues to exhibit the same 3 timestep 
offset that we observe with 3 covariates then we reason that the model must be learning across both temporal and feature dimensions simultaneously.
Figure \ref{fig:lstm_traffic_univariate_vs_covariate} illustrates the comparison between the univariate and covariate cases. 
It is evident from the plotted series, labelled as \textit{cov-2-pearsn-1.0-pl-8-seed-42-skip-1}, that the error closely resembles that of using 3 covariates.
\section{Discussion}
We have presented a repeatable method for testing the effectiveness of predicting covariates in neural networks that could be employed for alternative 
model architectures. Furthermore we have developed an LSTM model (seg-lstm) capable of producing state of the art results in a univariate setting on 
long forecast horizons without necessitating the need for additional features. 

Analysis of results has led us to reach the following conclusions:
\begin{itemize}
\item An LSTM is able to model both temporal and feature dynamics at short forecast horizons in the presence of short lead timestep covariates.
\item The effect covariates have on performance is, in part, related to the characteristics of the task (dataset) that the model is trained on.
\item The magnitude of the performance gain is related to the strength of the correlation between the covariate and the target variables and that this relationship 
is nonlinear with performance degrading most rapidly for $\pearson$ levels between 1.0 and 0.9. 
\item In most cases the presence of predicted covariates becomes either ineffective or a hindrance to performance as forecast horizons extend.
\item The number of timesteps covered by multiple covariates can offset the accumulation of error equal to the number of timesteps. 
\item The magnitude of the performance advantage from using covariates is related to the model's underlying ability to forecast accurately. 
Models that perform better in a univariate setting will produce comparatively better results in a 
covariate setting.
\item Increasing the number of covariates can enhance performance on short forecast horizons with the magnitude of the performance being related to the $\pearson$
\end{itemize}

Given the aim of this study was to explore the effect of covariates on LSTM networks, it's evident that predicting covariates 
jointly with target variables can under very limited conditions result in a performance improvement. As forecast horizons extend however, it frequently 
hinders performance. While acknowledging the artificial nature of the experiments used in this study
it nonetheless underscores some potential for covariates to assist neural networks in making more accurate predictions. 

It may be possible that the vanilla LSTM's struggle to learn the relationships between covariates and target variables 
across long forecast horizons and that alternative architectures and/or additional feature engineering techniques (such as LSTNet \cite{lai2018modeling} and the transformer based models
 of Zhou, Kitaev and Zhang \cite{zhou2021informer,kitaev2020reformer,zhang2023crossformer}) may perform better and a future study could evaluate their effectiveness using the same conditions. 

Future studies could also evaluate the use of covariates with more representative 
real-world data, such as the work by Zhou \cite{zhou2021informer} using the weather dataset which contains multiple features related to predicting weather. 
Additionally, further exploration with artificial datasets could investigate the effects of other characteristics like 
extending the covariate lead time offset on target variables, non-stationarity or negative correlations. This raises the question of whether providing 
the network with information regarding the temporal influence of each leading indicator on the target variable could simplify the learning task.

\bibliographystyle{unsrtnat}
\bibliography{rnn-covariates}  % replace 'your_bib_file_name' with the name of your .bib file, without the extension

%%%%%%%%%%%%%%%%%%%%%%%%%%%%%%%%%%%%%%%%%%%%%%%%%%%%%%%%%%%%

\FloatBarrier
\newpage
\appendix
\section{Appendix / supplemental material}
\subsection{Datasets}
All datasets used in this study are available through the Monash repository https://forecastingdata.org/ 
and are distributed under a Creative Commons Attribution 4.0 International licence.

\subsection{hyperparameters}
\begin{table}[htbp]
  \caption{hyperparameters}
  \label{hyperparameters}
  \centering
  \begin{tblr}{
    vline{-} = {1-11}{},
    hline{1-12} = {-}{},
  }
  learning rate           & 0.001    \\
  epochs                  & 100      \\
  hidden cells            & 40       \\
  lstm hidden layers      & 2        \\
  dropout                 & 0.1      \\
  weight decay            & 1e-8     \\
  batch size              & 128      \\
  batches per epoch       & 200 (base-lstm) 500 (seg-lstm)    \\
  early stopping patience & 30       \\
  optimizer               & AdamW    \\
  scheduler               & OneCycle \\
                          &          
  \end{tblr}
  \end{table}

\subsection{Computational Details}
All models were trained using CPU. \\
memory: 18 GB \\
os: macOS-10.16-x86\_64-i386-64bit \\
python 3.11 \\

\begin{table}[tbp]
  \caption{total model training times}
  \centering
  \begin{threeparttable}
  \begin{small}
  \renewcommand{\multirowsetup}{\centering}
  \setlength{\tabcolsep}{8pt}
  \begin{tabular}{lcc}
    \toprule
    \multirow{2}{*}{Dataset} & \multicolumn{2}{c}{Model} \\
    \cmidrule{2-3}
     & \textbf{base-lstm} (seconds) & \textbf{seg-lstm} (seconds)\\
    \midrule
    \multirow{1}{*}{\textbf{Electricity}} & $70442$ & $51839$ \\
    \midrule
    \multirow{1}{*}{\textbf{Hospital}} & $11029$ & $1151$ \\
    \midrule
    \multirow{1}{*}{\textbf{Tourism}} & $20412$ & $6807$ \\
    \midrule
    \multirow{1}{*}{\textbf{Traffic}} & $15667$ & $2652$ \\
    \bottomrule
  \end{tabular}
  \begin{tablenotes}
    \item[*] Each dataset is trained with 30 models.
  \end{tablenotes}
  \end{small}
  \end{threeparttable}
  \vspace{-15pt}
  \label{tab:training_time}
\end{table}

\subsection{Source Code}
The models were developed using Pytorch. Source code is available at \\
https://github.com/garethmd/nnts/tree/rnn-covariates-paper
\FloatBarrier

\subsection{Additional Results}
CSV Files of the full results are available on Google Drive \\
https://drive.google.com/drive/folders/179UK0lOOSi7yRdXOhcd7AFKl9U9A1DpE?usp=sharing
In this section we present additional tables and plots of results which supplement the results in the main paper. 

Table \ref{tab:base_lstm_covariate_mae_results} details
the MAE error for the base-lstm model for each dataset with various k covariates and correlation PCC values.  

\begin{table}[tbp]
  \caption{MAE base-lstm results for covariates $k \in \{0, 1, 2, 3\}$ and cross correlation $\pearson \in \{1, 0.9, 0.5\}$ values. }
  \centering
  \begin{threeparttable}
  \begin{small}
  \renewcommand{\multirowsetup}{\centering}
  \setlength{\tabcolsep}{1.8pt}
  \begin{tabular}{c|c|ccccccccccc}
    \toprule
    \multicolumn{2}{c}{Models} & \multicolumn{3}{c}{\textbf{base-lstm}} \\
    \cmidrule(lr){3-5} \cmidrule(lr){6-8}
    \multicolumn{2}{c}{$\pearson$} & 1 & 0.9 & 0.5 \\
    \toprule
    \multirow{4}{*}{\makecell{\textbf{Hospital}\\(12)}} & k=0 & 18.03 $\pm$ 0.306 & 18.03 $\pm$ 0.306 & 18.03 $\pm$ 0.306 \\
    & k=1 & 16.71 & 19.03 & 20.18 \\
    & k=2 & 15.62 & 20.31 & 20.77 \\
    & k=3 & 14.49 & 20.18 & 20.23 \\
    \midrule
    \multirow{4}{*}{\makecell{\textbf{Tourism}\\(24)}} & k=0 & 2336.42 $\pm$ 147.6 & 2336.42 $\pm$ 147.6 & 2336.42 $\pm$ 147.6 \\
    & k=1 & 2131.98 & 2545.43 & 3437.56 \\
    & k=2 & 2352.32 & 3734.32 & 4014.21 \\
    & k=3 & 3593.42 & 3340.92 & 3574.85 \\
    \midrule
    \multirow{4}{*}{\makecell{\textbf{Traffic}\\(8)}} & k=0 & 1.15 $\pm$ 0.010 & 1.15 $\pm$ 0.010 & 1.15 $\pm$ 0.010 \\
    & k=1 & 1.03 & 1.14 & 1.16 \\
    & k=2 & 0.95 & 1.09 & 1.14 \\
    & k=3 & 0.84 & 1.06 & 1.23 \\
    \midrule
    \multirow{4}{*}{\makecell{\textbf{Electricity}\\(168)}} & k=0 & 525.50 $\pm$ 51.74 & 525.50 $\pm$ 51.74 & 525.50 $\pm$ 51.74 \\
    & k=1 & 505.01 & 526.33 & 596.59 \\
    & k=2 & 528.35 & 452.38 & 519.68 \\ 
    & k=3 & 631.98 & 510.47 & 525.58 \\
    \bottomrule
  \end{tabular}
  \begin{tablenotes}
    \item Forecast horizons given in the brackets. Univariate k=0 intervals are 95\% CI.
  \end{tablenotes}
  \end{small}
  \end{threeparttable}
  \label{tab:base_lstm_covariate_mae_results}
  \vspace{-15pt}
\end{table}

Table \ref{tab:seg_lstm_covariate_mae_results} details
the MAE error for the seg-lstm model for each dataset with various k covariates and correlation PCC values.  

\begin{table}[tbp]
  \caption{MAE seg-lstm results for covariates $k \in \{0, 1, 2, 3\}$ and cross correlation $\pearson \in \{1, 0.9, 0.5\}$ values. }
  \centering
  \begin{threeparttable}
  \begin{small}
  \renewcommand{\multirowsetup}{\centering}
  \setlength{\tabcolsep}{1.8pt}
  \begin{tabular}{c|c|ccccccccccc}
    \toprule
    \multicolumn{2}{c}{Models} & \multicolumn{3}{c}{\textbf{seg-lstm}} \\
    \cmidrule(lr){3-5} \cmidrule(lr){6-8}
    \multicolumn{2}{c}{$\pearson$} & 1 & 0.9 & 0.5 \\
    \toprule
    \multirow{4}{*}{\makecell{\textbf{Hospital}\\(12)}} & k=0 & 19.95 $\pm$ 0.309 & 19.95 $\pm$ 0.309 & 19.95 $\pm$ 0.309 \\
    & k=1 & 16.73 & 20.83 & 21.55 \\
    & k=2 & 20.05 & 23.14 & 23.46 \\
    & k=3 & 21.34 & 22.22 & 24.93 \\
    \midrule
    \multirow{4}{*}{\makecell{\textbf{Tourism}\\(24)}} & k=0 & 1956.07 $\pm$ 163.4 & 1956.07 $\pm$ 163.4 & 1956.07 $\pm$ 163.4 \\
    & k=1 & 2078.31 & 3180.39 & 2292.99 \\
    & k=2 & 1782.48 & 3176.35 & 2857.37 \\
    & k=3 & 2053.37 & 2136.06 & 2790.61 \\
    \midrule
    \multirow{4}{*}{\makecell{\textbf{Traffic}\\(8)}} & k=0 & 1.17 $\pm$ 0.021 & 1.17 $\pm$ 0.021 & 1.17 $\pm$ 0.021 \\
    & k=1 & 1.04 & 1.16 & 1.17 \\
    & k=2 & 0.98 & 1.19 & 1.15 \\
    & k=3 & 0.87 & 1.07 & 1.16 \\
    \midrule
    \multirow{4}{*}{\makecell{\textbf{Electricity}\\(168)}} & k=0 & 287.95 $\pm$ 50.88 & 287.95 $\pm$ 50.88 & 287.95 $\pm$ 50.88 \\
    & k=1 & 286.77 & 305.21 & 290.34 \\
    & k=2 & 288.11 & 306.79 & 297.81\\ 
    & k=3 & 298.37 & 319.63 & 312.56 \\
    \bottomrule
  \end{tabular}
  \begin{tablenotes}
    \item Forecast horizons given in the brackets. Univariate k=0 intervals are 95\% CI.
  \end{tablenotes}
  \end{small}
  \end{threeparttable}
  \label{tab:seg_lstm_covariate_mae_results}
  \vspace{-15pt}
\end{table}

Table \ref{tab:base_lstm_covariate_rmse_results} details
the RMSE error for the base-lstm model for each dataset with various k covariates and correlation PCC values.  

\begin{table}[tbp]
  \caption{RMSE base-lstm results for covariates $k \in \{0, 1, 2, 3\}$ and cross correlation $\pearson \in \{1, 0.9, 0.5\}$ values. }
  \centering
  \begin{threeparttable}
  \begin{small}
  \renewcommand{\multirowsetup}{\centering}
  \setlength{\tabcolsep}{1.8pt}
  \begin{tabular}{c|c|ccccccccccc}
    \toprule
    \multicolumn{2}{c}{Models} & \multicolumn{3}{c}{\textbf{base-lstm}} \\
    \cmidrule(lr){3-5} \cmidrule(lr){6-8}
    \multicolumn{2}{c}{$\pearson$} & 1 & 0.9 & 0.5 \\
    \toprule
    \multirow{4}{*}{\makecell{\textbf{Hospital}\\(12)}} & k=0 & 22.03 $\pm$ 0.339 & 22.03 $\pm$ 0.339 & 22.03 $\pm$ 0.339  \\
    & k=1 & 20.95 & 23.22 & 24.53 \\
    & k=2 & 20.18 & 24.62 & 25.15 \\
    & k=3 & 19.65 & 24.51 & 24.55 \\
    \midrule
    \multirow{4}{*}{\makecell{\textbf{Tourism}\\(24)}} & k=0 & 2964.96 $\pm$ 155.8 & 2964.96 $\pm$ 155.8 & 2964.96 $\pm$ 155.8  \\
    & k=1 & 2764.97 & 3178.28 & 4409.48 \\
    & k=2 & 3116.47 & 4706.51 & 5015.75 \\
    & k=3 & 4575.04 & 4361.75 & 4730.95 \\
    \midrule
    \multirow{4}{*}{\makecell{\textbf{Traffic}\\(8)}} & k=0 & 1.56 $\pm$ 0.010 & 1.56 $\pm$ 0.010 & 1.56 $\pm$ 0.010  \\
    & k=1 & 1.47 & 1.55 & 1.58 \\
    & k=2 & 1.40 & 1.49 & 1.52 \\
    & k=3 & 1.28 & 1.46 & 1.68 \\
    \midrule
    \multirow{4}{*}{\makecell{\textbf{Electricity}\\(168)}} & k=0 & 675.03 $\pm$ 5.865 & 675.03 $\pm$ 5.865 & 675.03 $\pm$ 5.865 \\
    & k=1 & 650.48 & 689.42 & 734.66 \\
    & k=2 & 680.63 & 606.75 & 666.07 \\ 
    & k=3 & 783.84 & 675.23 & 668.15 \\
    \bottomrule
  \end{tabular}
  \begin{tablenotes}
    \item Forecast horizons given in the brackets. Univariate k=0 intervals are 95\% CI.
  \end{tablenotes}
  \end{small}
  \end{threeparttable}
  \label{tab:base_lstm_covariate_rmse_results}
  \vspace{-15pt}
\end{table}

Table \ref{tab:seg_lstm_covariate_rmse_results} details
the RMSE error for the seg-lstm model for each dataset with various k covariates and correlation PCC values.  
\begin{table}[tbp]
  \caption{RMSE seg-lstm results for covariates $k \in \{0, 1, 2, 3\}$ and cross correlation $\pearson \in \{1, 0.9, 0.5\}$ values. }
  \centering
  \begin{threeparttable}
  \begin{small}
  \renewcommand{\multirowsetup}{\centering}
  \setlength{\tabcolsep}{1.8pt}
  \begin{tabular}{c|c|ccccccccccc}
    \toprule
    \multicolumn{2}{c}{Models} & \multicolumn{3}{c}{\textbf{seg-lstm}} \\
    \cmidrule(lr){3-5} \cmidrule(lr){6-8}
    \multicolumn{2}{c}{$\pearson$} & 1 & 0.9 & 0.5 \\
    \toprule
    \multirow{4}{*}{\makecell{\textbf{Hospital}\\(12)}} & k=0 & 24.19 $\pm$ 0.434 & 24.19 $\pm$ 0.434 & 24.19 $\pm$ 0.434 \\
    & k=1 & 20.99 & 25.47 & 26.41 \\
    & k=2 & 24.55 & 27.93 & 28.27 \\
    & k=3 & 26.12 & 26.92 & 30.01 \\
    \midrule
    \multirow{4}{*}{\makecell{\textbf{Tourism}\\(24)}} & k=0 & 2964.96 $\pm$ 155.8 & 2964.96 $\pm$ 155.8 & 2964.96 $\pm$ 155.8 \\
    & k=1 & 2612.12 & 3803.94 & 2833.16 \\
    & k=2 & 2228.58 & 3875.71 & 3497.33 \\
    & k=3 & 2647.94 & 2660.67 & 3467.14 \\
    \midrule
    \multirow{4}{*}{\makecell{\textbf{Traffic}\\(8)}} & k=0 & 1.56 $\pm$ 0.010 & 1.56 $\pm$ 0.010 & 1.56 $\pm$ 0.010 \\
    & k=1 & 1.48 & 1.57 & 1.57 \\
    & k=2 & 1.44 & 1.61 & 1.55 \\
    & k=3 & 1.32 & 1.45 & 1.53 \\
    \midrule
    \multirow{4}{*}{\makecell{\textbf{Electricity}\\(168)}} & k=0 & 469.07 $\pm$ 7.734 & 469.07 $\pm$ 7.734 & 469.07 $\pm$ 7.734 \\
    & k=1 & 481.60 & 482.13 & 472.95 \\
    & k=2 & 474.75 & 488.84 & 473.14 \\ 
    & k=3 & 481.93 & 500.84 & 492.91 \\
    \bottomrule
  \end{tabular}
  \begin{tablenotes}
    \item Forecast horizons given in the brackets. Univariate k=0 intervals are 95\% CI.
  \end{tablenotes}
  \end{small}
  \end{threeparttable}
  \label{tab:seg_lstm_covariate_rmse_results}
  \vspace{-15pt}
\end{table}
\FloatBarrier

\begin{figure}[ht]

\begin{subfigure}{\textwidth}
\centering
\includegraphics[width=\textwidth]{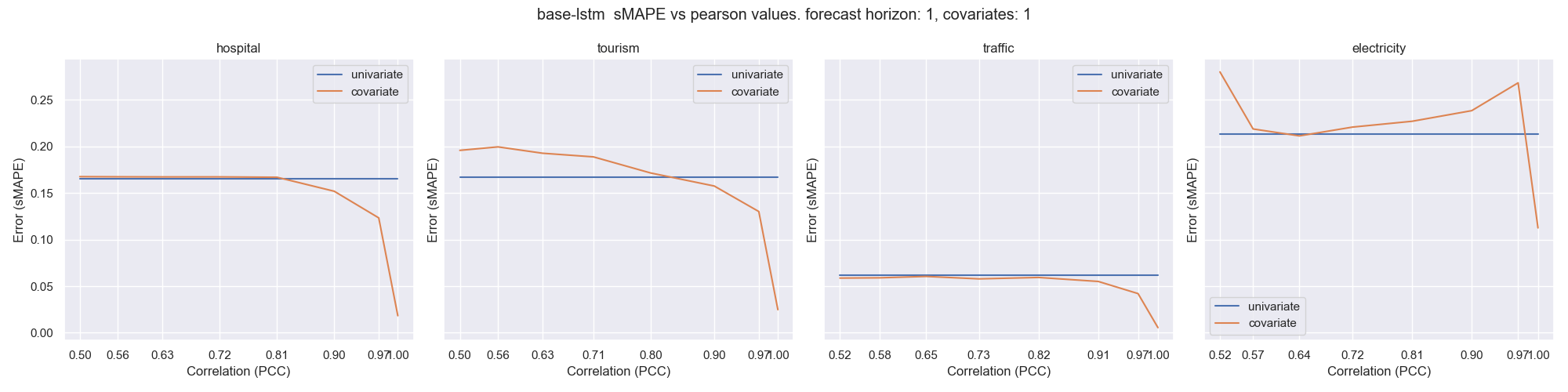}
\caption{base-lstm smape as a function of PCC for 1 covariate}
\label{fig:base_lstm_k_1_smape_vs_pearson}
\end{subfigure}

\begin{subfigure}{\textwidth}
\centering
\includegraphics[width=\textwidth]{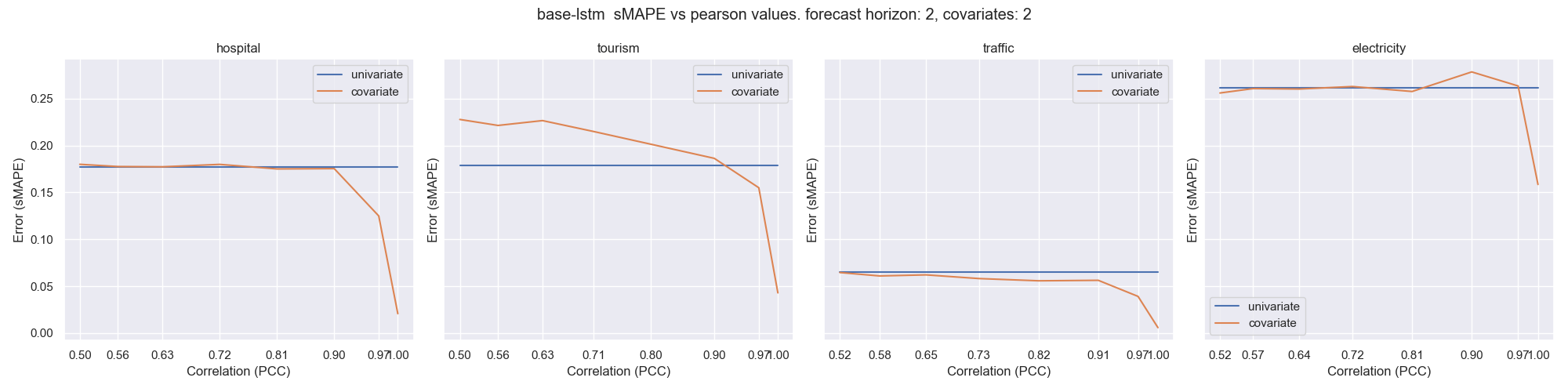}
\caption{base-lstm smape as a function of PCC for 2 covariates}
\label{fig:base_lstm_k_2_smape_vs_pearson}
\end{subfigure}

\begin{subfigure}{\textwidth}
\centering
\includegraphics[width=\textwidth]{figures/base_k3_pe_pcc.png}
\caption{base-lstm smape as a function of PCC for 3 covariates}
\label{fig:appendix_base_lstm_k_3_smape_vs_pearson}
\end{subfigure}
\caption{base-lstm sMAPE for various covariates as a function of correlation (PCC)}
\label{fig:base_lstm_smape_vs_pearson}
\end{figure}

Figures \ref{fig:base_lstm_smape_vs_pearson} plots the sMAPE as a function of correlation (PCC) for base-lstm models

\begin{figure}[ht]
\begin{subfigure}{\textwidth}
\centering
\includegraphics[width=\textwidth]{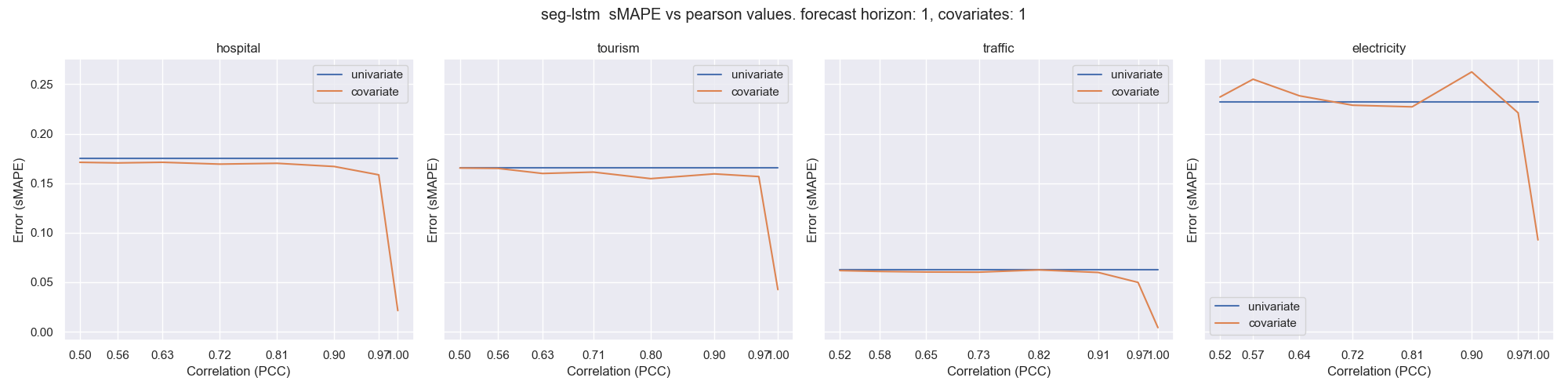}
\caption{seg-lstm smape as a function of PCC for 1 covariate}
\label{fig:seg_lstm_k_1_smape_vs_pearson}
\end{subfigure}
\begin{subfigure}{\textwidth}
\centering
\includegraphics[width=\textwidth]{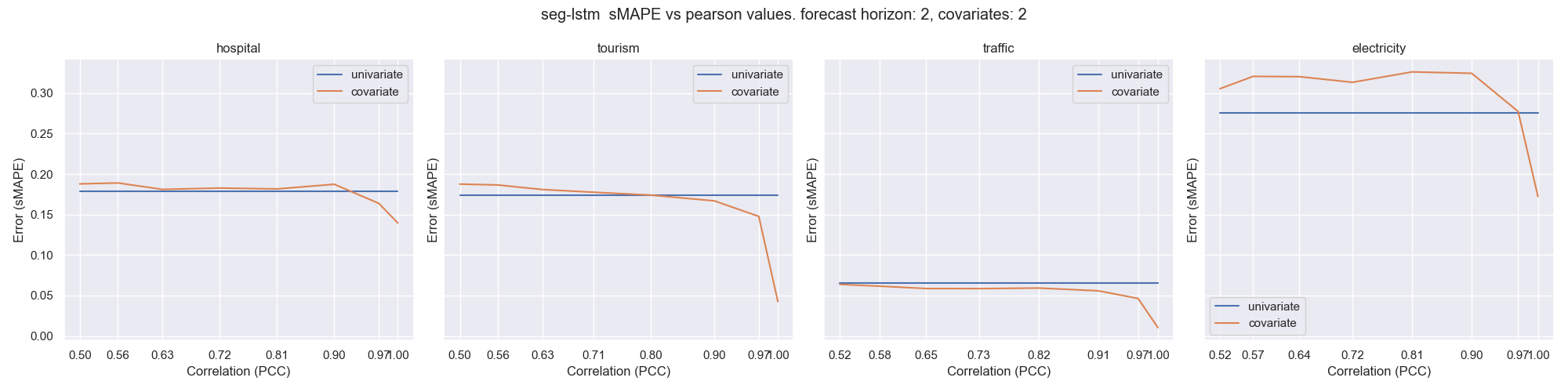}
\caption{seg-lstm smape as a function of PCC for 2 covariates}
\label{fig:seg_lstm_k_2_smape_vs_pearson}
\end{subfigure}
\begin{subfigure}{\textwidth}
\centering
\includegraphics[width=\textwidth]{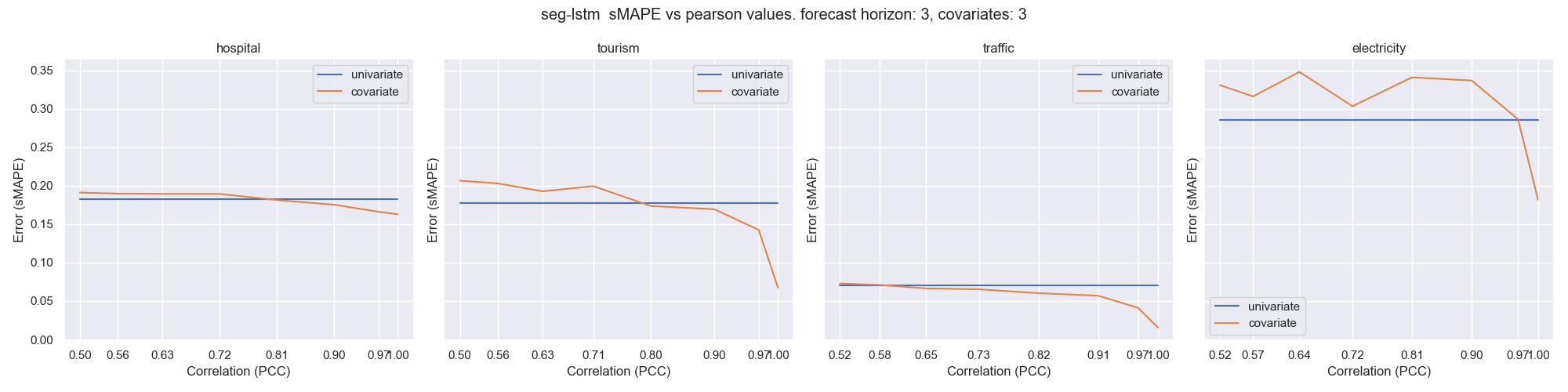}
\caption{seg-lstm smape as a function of PCC for 3 covariates}
\label{fig:seg_lstm_k_3_smape_vs_pearson}
\end{subfigure}
\caption{seg-lstm sMAPE for various covariates as a function of correlation (PCC)}
\label{fig:seg_lstm_smape_vs_pearson}
\end{figure}

Figures \ref{fig:seg_lstm_smape_vs_pearson} plots the sMAPE as a function of correlation (PCC) for seg-lstm models

\begin{figure}[tbp]
\begin{subfigure}{\textwidth}
\centering
\includegraphics[width=\linewidth]{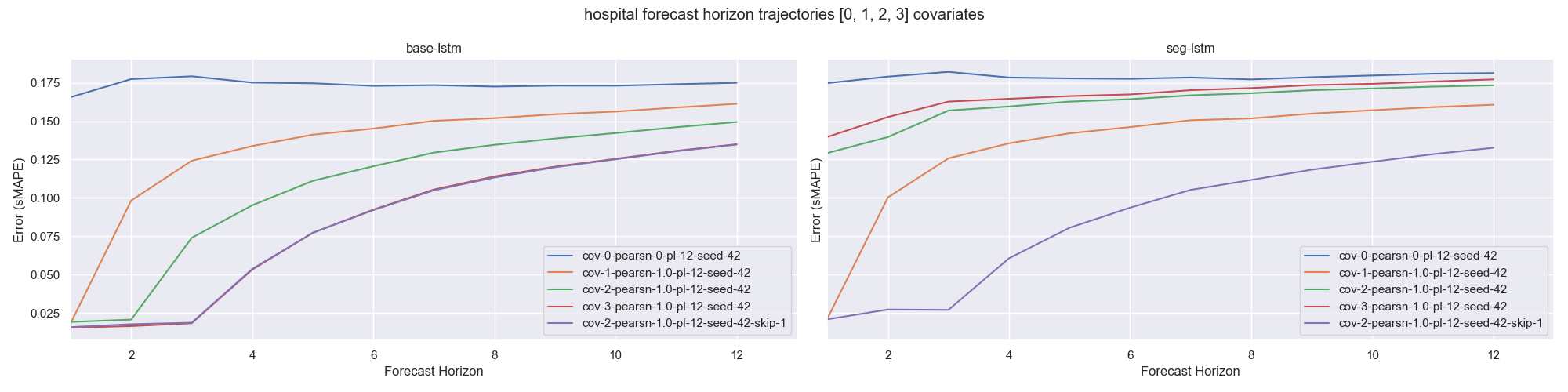}
\caption{hospital forecast horizon trajectories}
\label{fig:base-lstm_seg-lstm_hospital_k_0_1_2_3_trajectory}
\end{subfigure}

\begin{subfigure}{\textwidth}
\centering
\includegraphics[width=\linewidth]{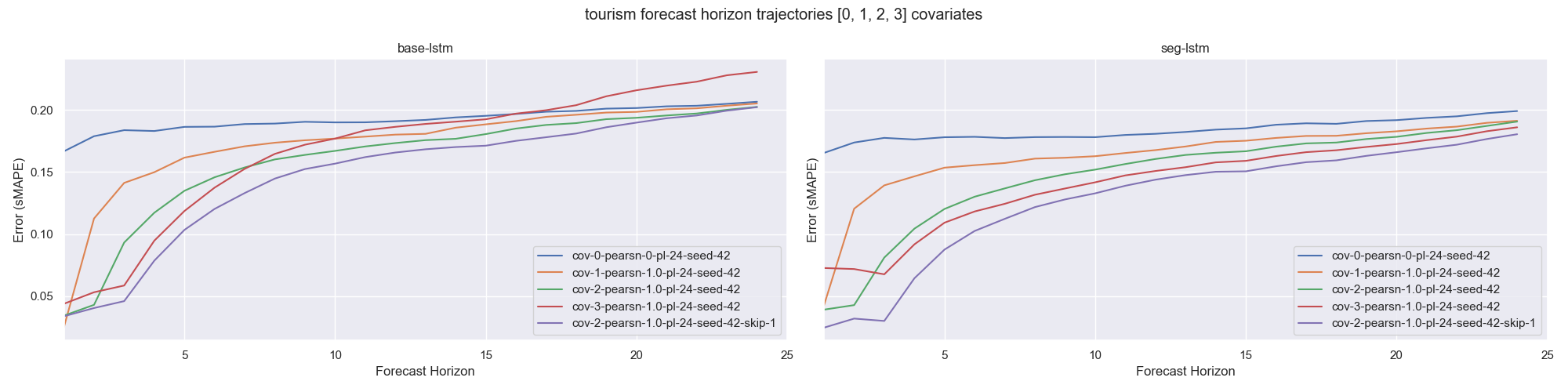}
\caption{tourism forecast horizon trajectories}
\label{fig:base-lstm_seg-lstm_tourism_k_0_1_2_3_trajectory}
\end{subfigure}

\begin{subfigure}{\textwidth}
\centering
\includegraphics[width=\linewidth]{figures/models_traf_k.png}
\caption{traffic forecast horizon trajectories}
\label{fig:base-lstm_seg-lstm_traffic_k_0_1_2_3_trajectory}
\end{subfigure}

\begin{subfigure}{\textwidth}
\centering
\includegraphics[width=\linewidth]{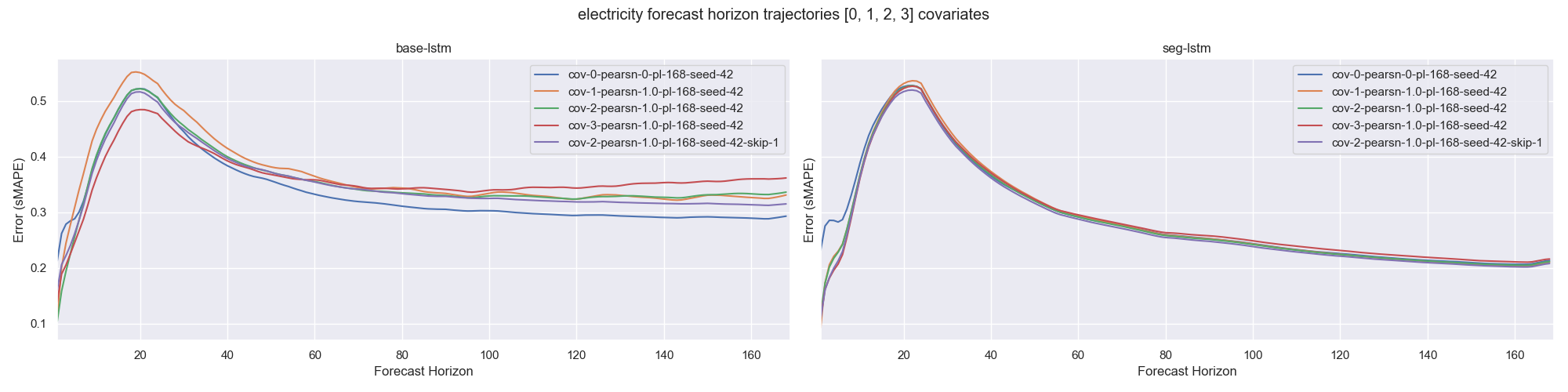}
\caption{electricity forecast horizon trajectories}
\label{fig:base-lstm_seg-lstm_electricity_k_0_1_2_3_trajectory}
\end{subfigure}

\caption{sMAPE for various covariates as a function of Forecast Horizon for correlation (PCC) = 1.0}
\label{fig:base-lstm_seg-lstm_k_0_1_2_3_trajectory}
\end{figure}

Figures \ref{fig:base-lstm_seg-lstm_k_0_1_2_3_trajectory} plots sMAPE for various covariates as a function of Forecast Horizon for correlation (PCC) = 1.0

\begin{figure}[tbp]
\begin{subfigure}{\textwidth}
\centering
\includegraphics[width=\linewidth]{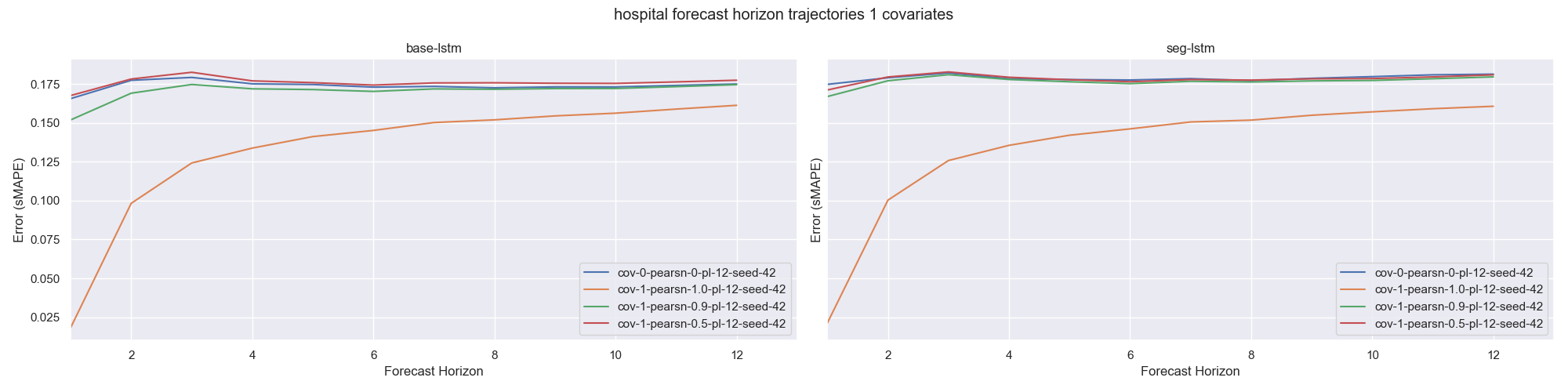}
\caption{hospital forecast horizon trajectories k=1}
\label{fig:base-lstm_seg-lstm_hospital_k_1_trajectory}
\end{subfigure}

\begin{subfigure}{\textwidth}
\centering
\includegraphics[width=\linewidth]{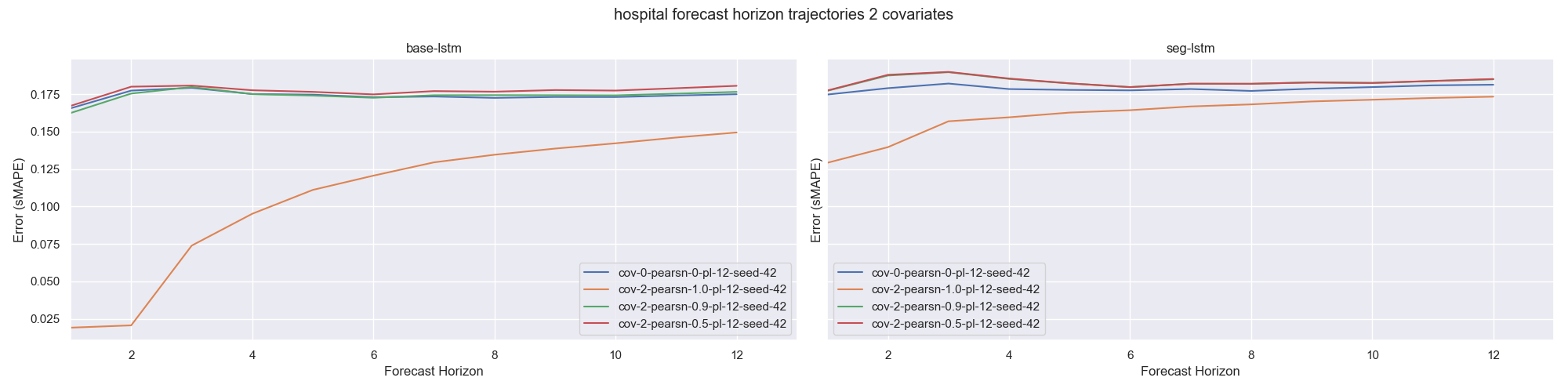}
\caption{hospital forecast horizon trajectories k=2}
\label{fig:base-lstm_seg-lstm_hospital_k_2_trajectory}
\end{subfigure}

\begin{subfigure}{\textwidth}
\centering
\includegraphics[width=\linewidth]{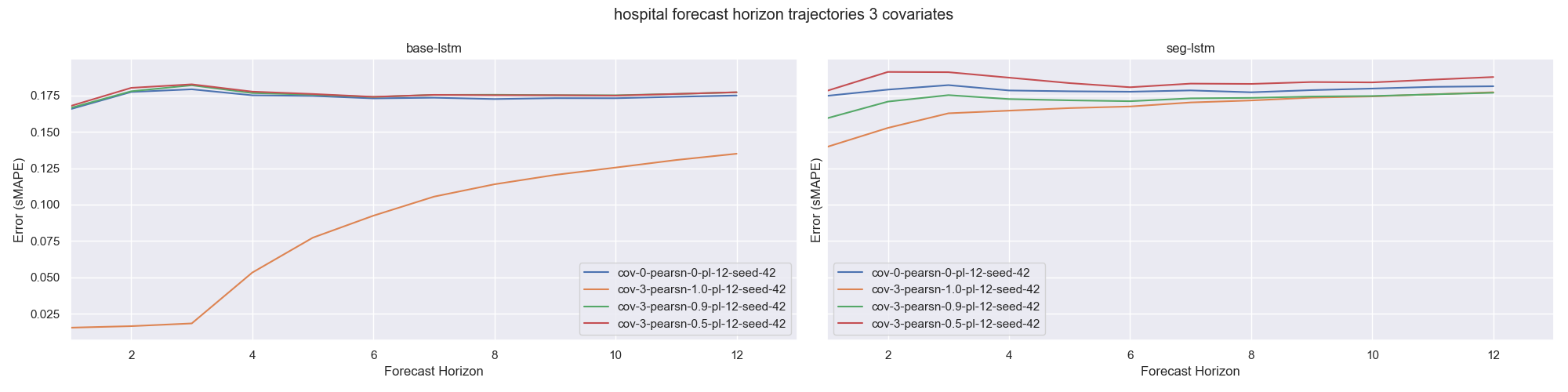}
\caption{hospital forecast horizon trajectories k=3}
\label{fig:base-lstm_seg-lstm_hospital_k_3_trajectory}
\end{subfigure}
\caption{hospital forecast horizon trajectories for various covariates k and correlation values (PCC)}
\label{fig:base-lstm_seg-lstm_hospital_trajectory}
\end{figure}

Figures \ref{fig:base-lstm_seg-lstm_hospital_trajectory} plots sMAPE hospital forecast horizon trajectories for various covariates k and correlation values (PCC)

\begin{figure}[tbp]

\begin{subfigure}{\textwidth}
\centering
\includegraphics[width=\linewidth]{figures/models_tour_k1.png}
\caption{tourism forecast horizon trajectories k=1}
\label{fig:base-lstm_seg-lstm_tourism_k_1_trajectory}
\end{subfigure}

\begin{subfigure}{\textwidth}
\centering
\includegraphics[width=\linewidth]{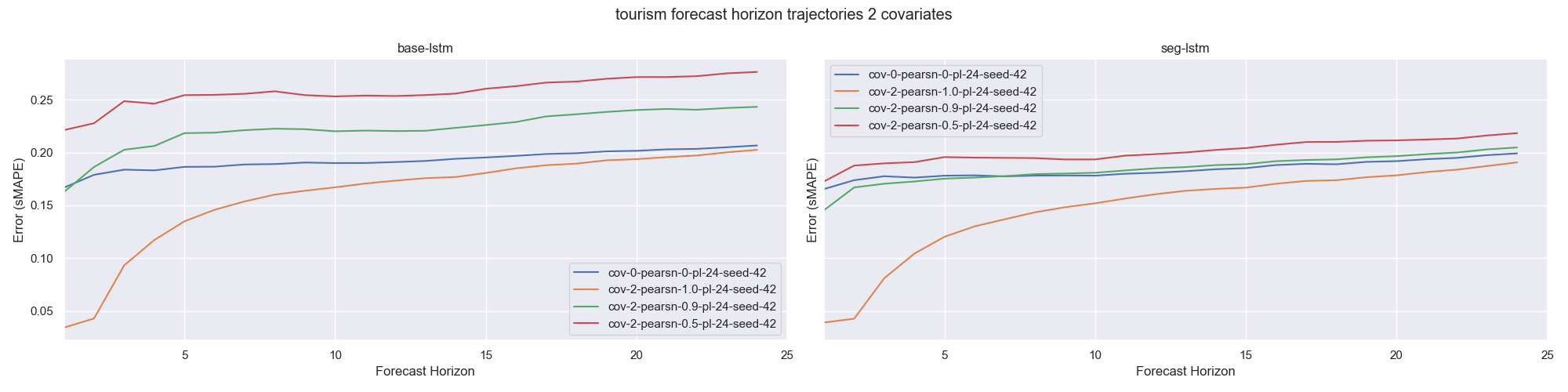}
\caption{tourism forecast horizon trajectories k=2}
\label{fig:base-lstm_seg-lstm_tourism_k_2_trajectory}
\end{subfigure}

\begin{subfigure}{\textwidth}
\centering
\includegraphics[width=\linewidth]{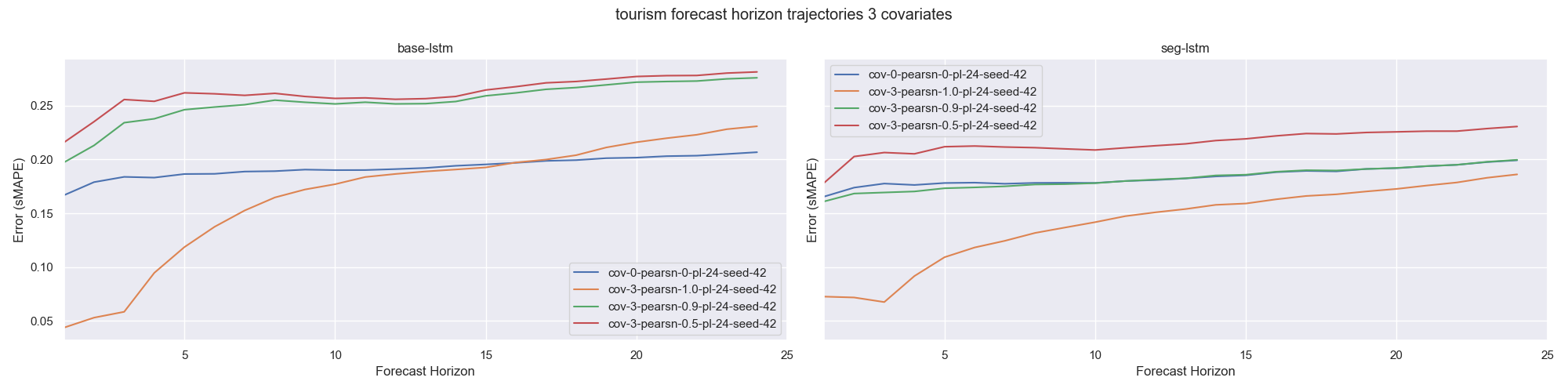}
\caption{tourism forecast horizon trajectories k=3}
\label{fig:base-lstm_seg-lstm_tourism_k_3_trajectory}
\end{subfigure}
\caption{tourism forecast horizon trajectories for various covariates k and correlation values (PCC)}
\label{fig:base-lstm_seg-lstm_tourism_trajectory}
\end{figure}

Figures \ref{fig:base-lstm_seg-lstm_tourism_trajectory} plots sMAPE tourism forecast horizon trajectories for various covariates k and correlation values (PCC)

\begin{figure}[tbp]
\begin{subfigure}{\textwidth}
\centering
\includegraphics[width=\linewidth]{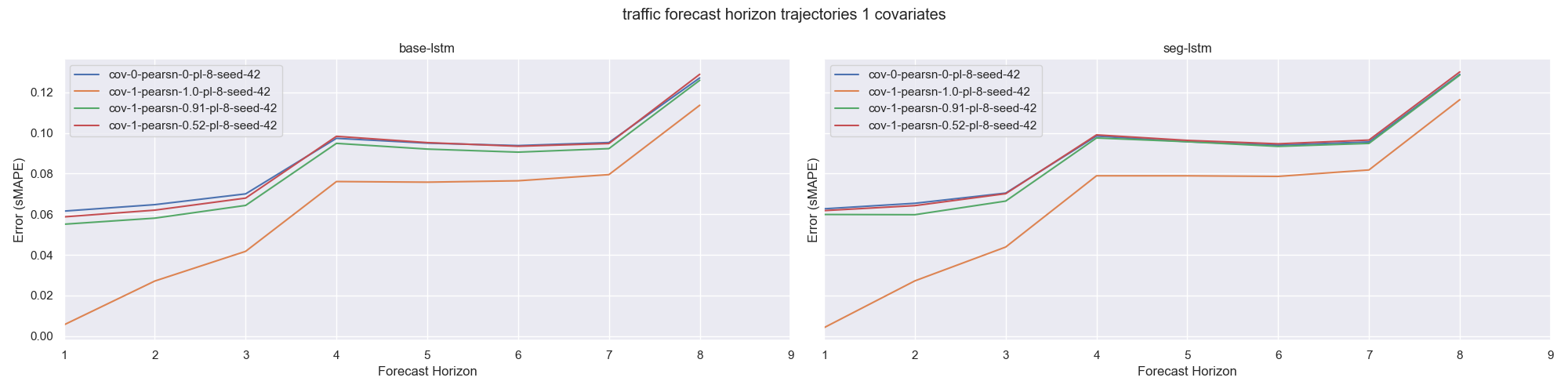}
\caption{traffic forecast horizon trajectories k=1}
\label{fig:base-lstm_seg-lstm_traffic_k_1_trajectory}
\end{subfigure}

\begin{subfigure}{\textwidth}
\centering
\includegraphics[width=\linewidth]{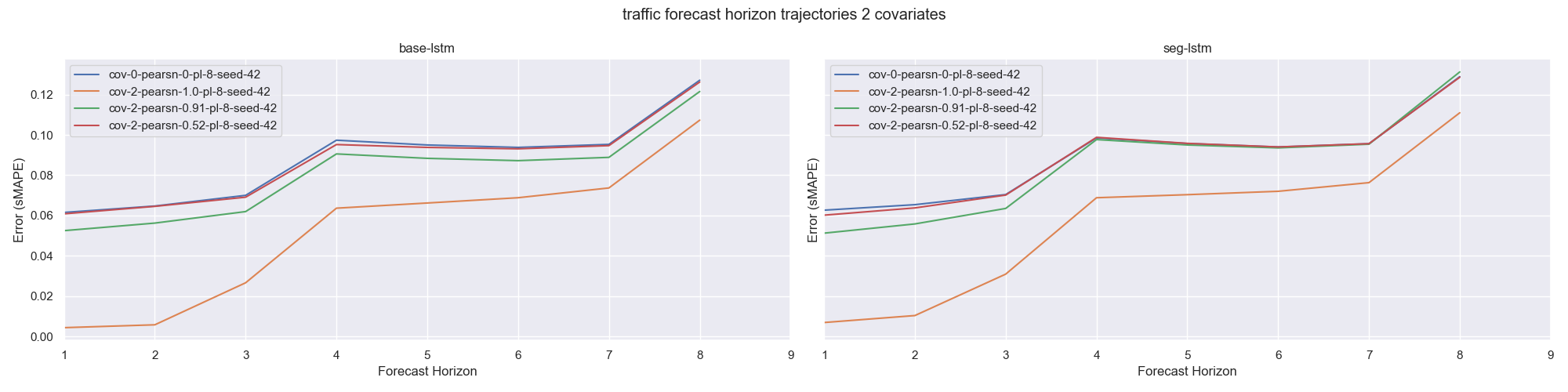}
\caption{traffic forecast horizon trajectories k=2}
\label{fig:base-lstm_seg-lstm_traffic_k_2_trajectory}
\end{subfigure}

\begin{subfigure}{\textwidth}
\centering
\includegraphics[width=\linewidth]{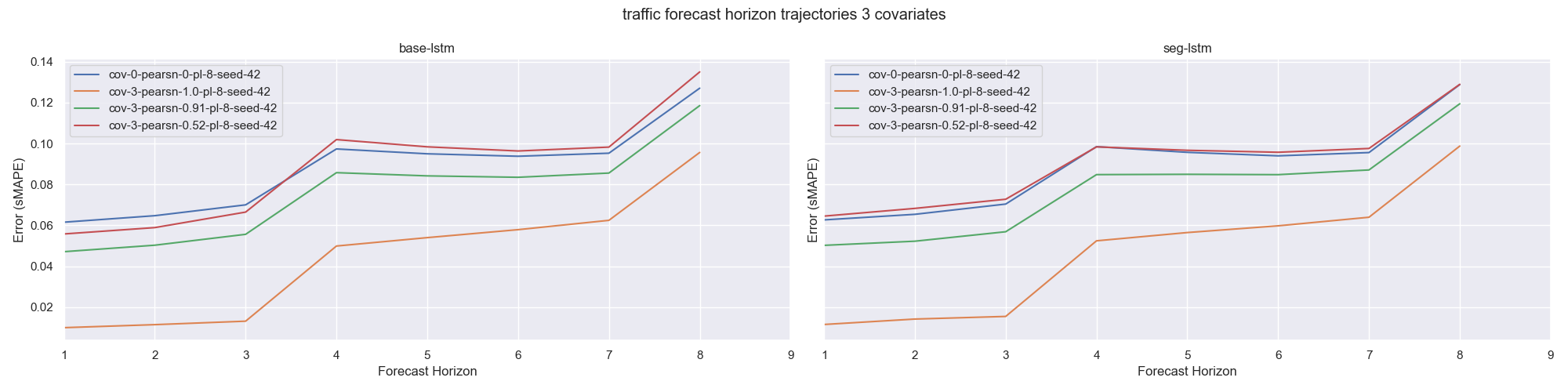}
\caption{traffic forecast horizon trajectories k=3}
\label{fig:base-lstm_seg-lstm_traffic_k_3_trajectory}
\end{subfigure}
\caption{traffic forecast horizon trajectories for various covariates k and correlation values (PCC)}
\label{fig:base-lstm_seg-lstm_traffic_trajectory}
\end{figure}

Figures \ref{fig:base-lstm_seg-lstm_traffic_trajectory} plots sMAPE traffic forecast horizon trajectories for various covariates k and correlation values (PCC)

\begin{figure}[tbp]

\begin{subfigure}{\textwidth}
\centering
\includegraphics[width=\linewidth]{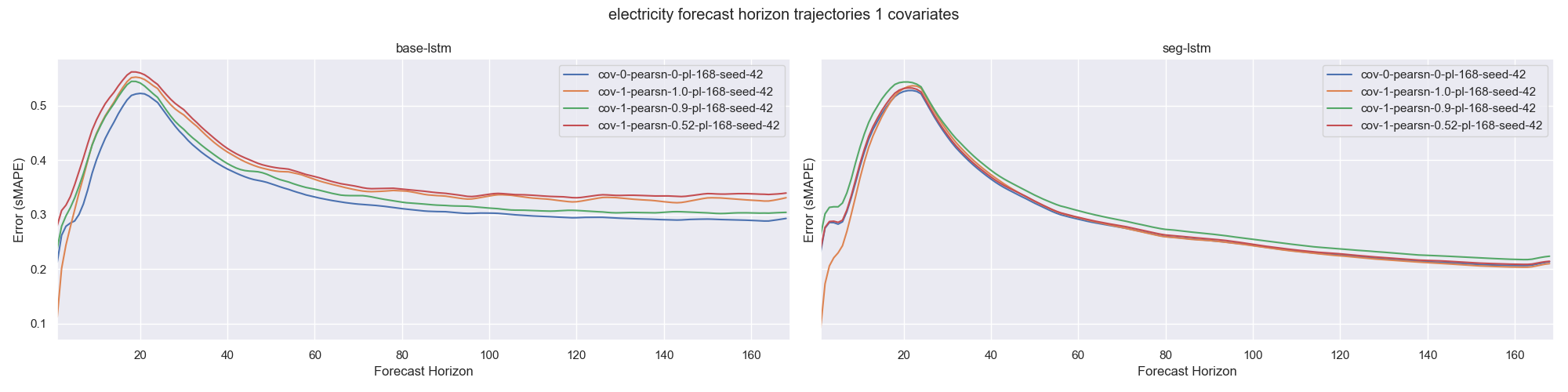}
\caption{electricity forecast horizon trajectories k=1}
\label{fig:base-lstm_seg-lstm_electricity_k_1_trajectory}
\end{subfigure}

\begin{subfigure}{\textwidth}
\centering
\includegraphics[width=\linewidth]{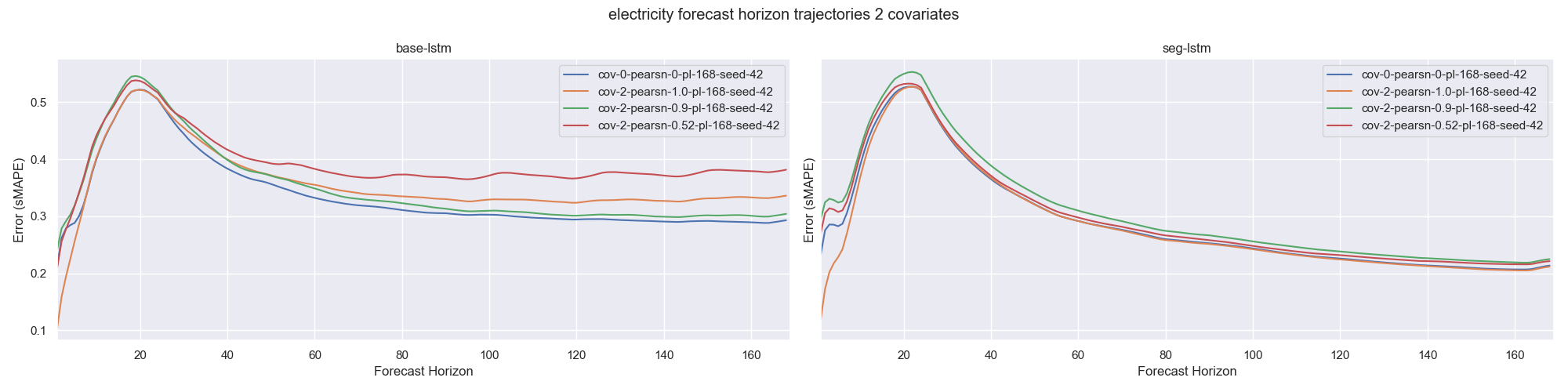}
\caption{electricity forecast horizon trajectories k=2}
\label{fig:base-lstm_seg-lstm_electricity_k_2_trajectory}
\end{subfigure}

\begin{subfigure}{\textwidth}
\centering
\includegraphics[width=\linewidth]{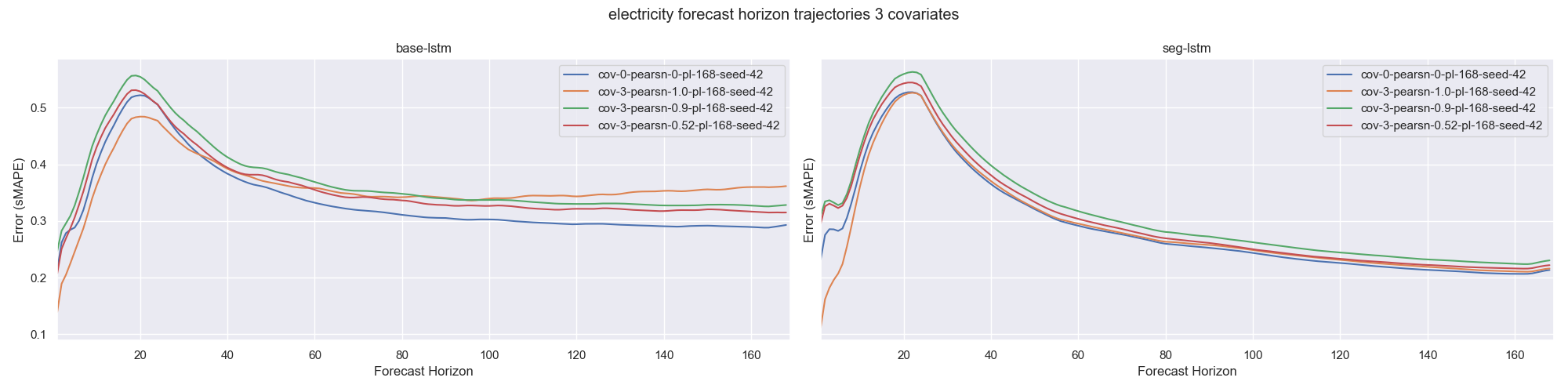}
\caption{electricity forecast horizon trajectories k=3}
\label{fig:base-lstm_seg-lstm_electricity_k_3_trajectory}
\end{subfigure}
\caption{electricity forecast horizon trajectories for various covariates k and correlation values (PCC)}
\label{fig:base-lstm_seg-lstm_electricity_trajectory}
\end{figure}

Figures \ref{fig:base-lstm_seg-lstm_electricity_trajectory} plos sMAPE electricity forecast horizon trajectories for various covariates k and correlation values (PCC)

\end{document}